\pdfoutput=1

\documentclass[11pt]{article}

\usepackage[final]{ACL2023}
\usepackage{longtable}
\usepackage{caption}
\usepackage{placeins}
\usepackage{url}
\usepackage{stfloats}

\newcommand\blfootnote[1]{%
  \begingroup
  \renewcommand\thefootnote{}\footnote{#1}%
  \addtocounter{footnote}{-1}%
  \endgroup
}

\usepackage{etoolbox}
\patchcmd{\footnotetext}{\indent}{}{}{} 
\usepackage{ragged2e} 

\usepackage{times}
\usepackage{latexsym}

\usepackage[T1]{fontenc}

\usepackage[utf8]{inputenc}

\usepackage{microtype}

\usepackage{inconsolata}
\usepackage{graphicx}

\title{Advancing Uto-Aztecan Language Technologies: \\ A Case Study on the Endangered Comanche Language
\\[10pt]  
\includegraphics[width=1.5cm]{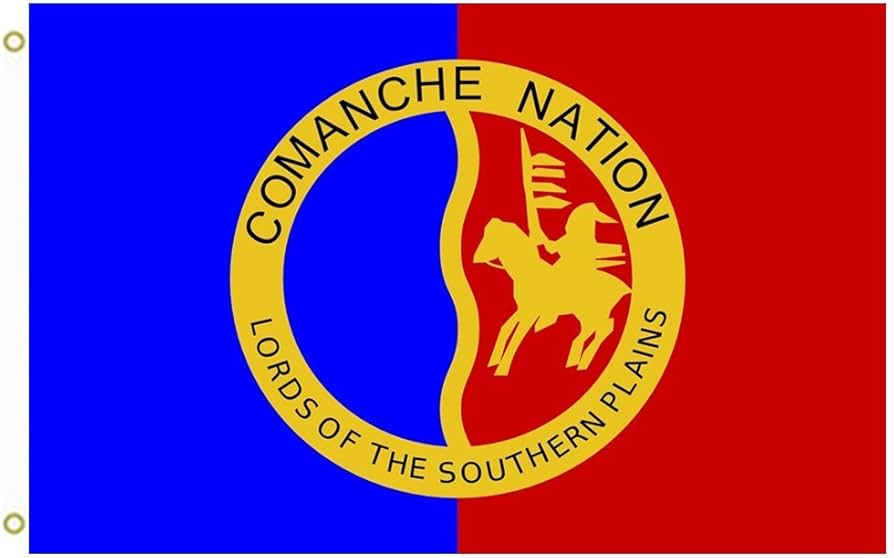}}

\author{
  Jesus Alvarez C,
  Daua D. Karajeanes,
  Ashley Celeste Prado,
  John Ruttan,\\
  \textbf{Ivory Yang},
  \textbf{Sean O'Brien},
  \textbf{Vasu Sharma},
  \textbf{Kevin Zhu} \\
  Algoverse AI Research \\
    \texttt{ivory.yang.gr@dartmouth.edu, kevin@algoverse.us}
}

\begin{document}
\maketitle

\blfootnote{
\small
\RaggedRight
Contact other authors at: jalvarezc@my.canyons.edu, d.d.karajeanes@student.tue.nl, aprad054@fiu.edu, jruttan3@uwo.ca, seobrien@ucsd.edu, vasus@andrew.cmu.edu
\par
}

\begin{abstract}
The digital exclusion of endangered languages remains a critical challenge in NLP, limiting both linguistic research and revitalization efforts. This study introduces the first computational investigation of Comanche, an Uto-Aztecan language on the verge of extinction, demonstrating how minimal-cost, community-informed NLP interventions can support language preservation. We present a manually curated dataset of 412 phrases, a synthetic data generation pipeline, and an empirical evaluation of GPT-4o and GPT-4o-mini for language identification. Our experiments reveal that while LLMs struggle with Comanche in zero-shot settings, few-shot prompting significantly improves performance, achieving near-perfect accuracy with just five examples. Our findings highlight the potential of targeted NLP methodologies in low-resource contexts and emphasize that visibility is the first step toward inclusion. By establishing a foundation for Comanche in NLP, we advocate for computational approaches that prioritize accessibility, cultural sensitivity, and community engagement.
\end{abstract}

\section{Introduction}
The decline of endangered languages represents not only a linguistic loss \cite{low2022endangered} but also the erosion of invaluable cultural, historical, and ecological knowledge \cite{Tulloch2006, camara2021language, sallabank2023endangered}. Despite growing advancements in language technology, computational efforts overwhelmingly favor widely spoken languages, leaving endangered languages largely unsupported \cite{meighan2021decolonizing, yang2025nushurescue, jerpelea2025dialectal}. Over 88\% of the world's languages have minimal to no representation in mainstream language technologies, exacerbating their digital marginalization \cite{Rangel2019}. This exclusion hinders linguistic research and deepens the digital divide \cite{valijarvi2023role, yang2025navajo}, complicating preservation and revitalization efforts. Among these, Comanche, an Uto-Aztecan language, faces imminent extinction, with fewer than 50 fluent speakers remaining \cite{Chaika2024}. 

\begin{figure}[t]
  \centering
  \includegraphics[width=\linewidth]{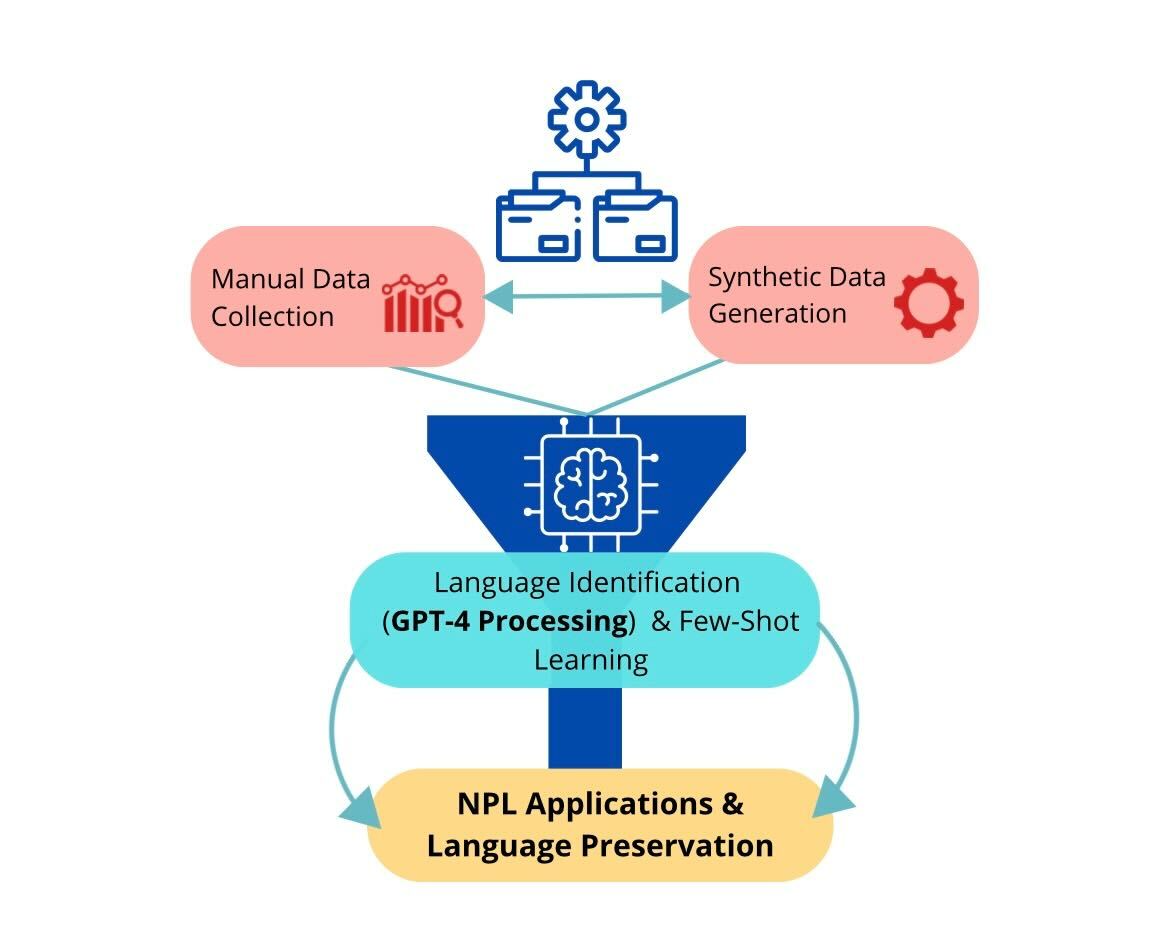}
  \caption{Stylized overview of our exploration of NLP applications for the endangered Comanche language.}
  \label{fig:intro}
\end{figure}


We present a case study on the Comanche language, demonstrating that with minimal cost and computational resources, it is possible to achieve what large corporations and academic institutions have largely neglected. As shown in Figure \ref{fig:intro}, we contribute (1) a manually curated dataset, (2) synthetic data generation pipeline for resource expansion, and (3) an empirical evaluation of GPT-4o and GPT-4o-mini in zero-shot and few-shot language identification. Our findings highlight the potential of large language models (LLMs) in low-resource settings, offering insights into their applicability for endangered language preservation. \textbf{This work marks the first-ever introduction of Comanche into the NLP domain, laying groundwork for future research and linguistic equity}.

\section{Related Work}

Efforts to preserve endangered languages, particularly Native American languages, date back to at least the early 20th century \cite{charney1993}, with early approaches relying heavily on linguistic documentation and literary preservation \cite{schwartz2016cultures}. While these foundational efforts paved the way, they were hindered by the scarcity of available datasets and standardized benchmarks, leading researchers to explore alternative strategies for text processing \cite{lorenzo2024mitigating, spencer2025can}. Despite these advancements, modern computational linguistics continues to face significant challenges when working with polysynthetic languages such as Comanche and Apache, due to their intricate orthographic and morphological structures \cite{kelly2020evaluation}. In recent years there have been promising community-led revitalization initiatives, including immersive education programs, digital archives, and collaborations with computational linguists \cite{bia2023, Schwartz2021}.

Data scarcity remains a fundamental challenge in NLP \cite{glaser2021data}. Unlike widely spoken languages with abundant corpora, low-resource languages lack annotated datasets, limiting the effectiveness of LLMs for preservation \cite{zhong2024opportunities, dinh2024multi}. Few-shot prompting has emerged as a promising solution, allowing LLMs to generate synthetic data from minimal examples \cite{zhang2021differentiable}, though its success hinges on data quality. Transfer learning \cite{adimulam2022transfer} has also been explored to improve low-resource NLP, but without robust evaluation frameworks tailored for Indigenous languages \cite{shu2024transcending}, achieving meaningful generalization remains a challenge \cite{Mager2023}.

\begin{figure}[t]
  \centering
  \includegraphics[width=\linewidth]{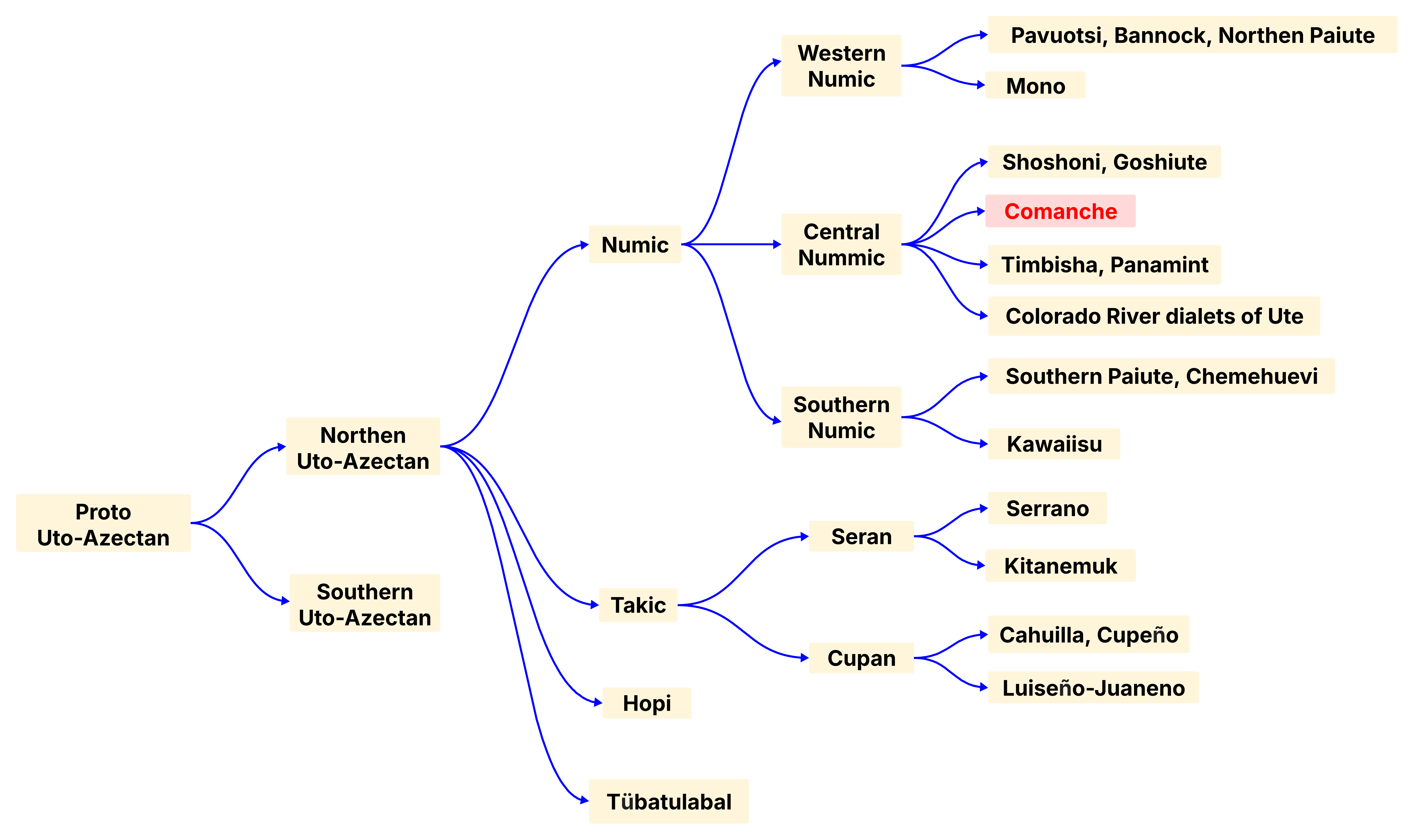}
  \caption{Family tree for Uto-Aztecan Languages, with Comanche highlighted.}
  \label{fig:familytree}
\end{figure}

\begin{figure*}[t]
  \centering
  \includegraphics[width=\linewidth]{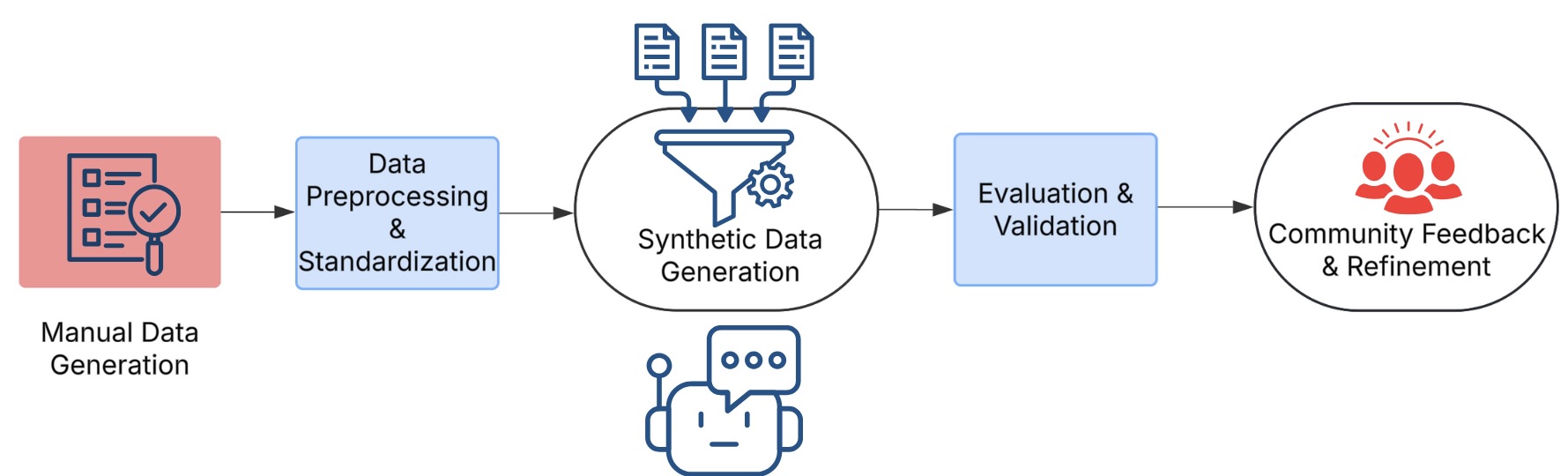}
  \caption{Data pipeline.}
  \label{datapipeline}
\end{figure*}

\section{Native American Language Landscape}
The linguistic diversity of Native American languages is vast, spanning multiple language families with distinct phonetic, morphological, and syntactic properties. Despite this richness, many of these languages are critically endangered, with fluency declining due to historical policies of forced assimilation, boarding schools, and sociopolitical marginalization \cite{krauss1992world}. Language documentation efforts have attempted to counteract this loss, but computational resources remain scarce, and mainstream NLP models are ill-equipped to process these languages effectively \cite{blasi-etal-2022-systematic}. The lack of digital resources further exacerbates the challenge, preventing these languages from benefiting from advances in language technologies \cite{DOI2022}.

Comanche belongs to the Uto-Aztecan language family, one of the largest language families in the Americas, encompassing over 60 languages spoken across the western United States, Mexico, and Central America \cite{opler1943origins}. While some Uto-Aztecan languages, such as Nahuatl \cite{andrews2003introduction}, have relatively larger speaker populations and a degree of digital presence, others, including Comanche, face imminent extinction. As shown in Figure \ref{fig:familytree}, Comanche developed as a distinct language after diverging from Shoshone in the 18th century, evolving unique phonological and lexical features \cite{casagrande1955comanche}. Today, with fewer than 50 fluent speakers, Comanche lacks sufficient linguistic resources for computational modeling.

\section{Data}
\subsection{Manual Data Collection}
To construct a foundational dataset for Comanche, we conducted a systematic review of linguistic resources, including academic literature, digital archives, and historical records. Given the scarcity of publicly available corpora, we aggregated and curated data from 15 distinct domains (Appendix \ref{sec:appendixb}), ensuring consistency through transcription and standardization. To enhance data reliability, we cross-referenced linguistic materials with community-driven documentation efforts, validating authenticity and linguistic accuracy. This structured dataset of 412 Comanche phrases, the first digitalized dataset of its kind, serves as a crucial resource for both language preservation and computational linguistic research in Comanche.

\subsection{Synthetic Data Generation}
Given the extreme scarcity of parallel Comanche–English text, we leveraged few-shot prompting with GPT-4o to generate synthetic translations. Using a manually curated dataset of 100 Comanche–English sentence pairs, we split the data into an 80\% training set and a 20\% test set. During training, GPT-4o was provided examples from the training subset and then prompted to generate translations for the test set (Appendix \ref{sec:appendixe}). The generated outputs were evaluated using normalized Levenshtein similarity, ensuring a minimum quality threshold of 0.1\footnote{Given that Comanche has never been explored in NLP, we set a baseline threshold of 0.1 due to the difficulty of the task. As the pipeline matures, we will refine our evaluation criteria and increase the required similarity score.} before incorporation into the dataset. This controlled expansion strategy maintained linguistic integrity while demonstrating that even minimal data can be effectively leveraged to create valuable resources for endangered language NLP. While the pipeline shown in Figure \ref{datapipeline} is in early stage, it underscores the potential of leveraging NLP for endangered language documentation and expansion. As data scarcity persists, synthetic augmentation offers a scalable approach to bridge resource gaps and support revitalization efforts.

\begin{figure}[t]
  \centering
  \includegraphics[width=\linewidth]{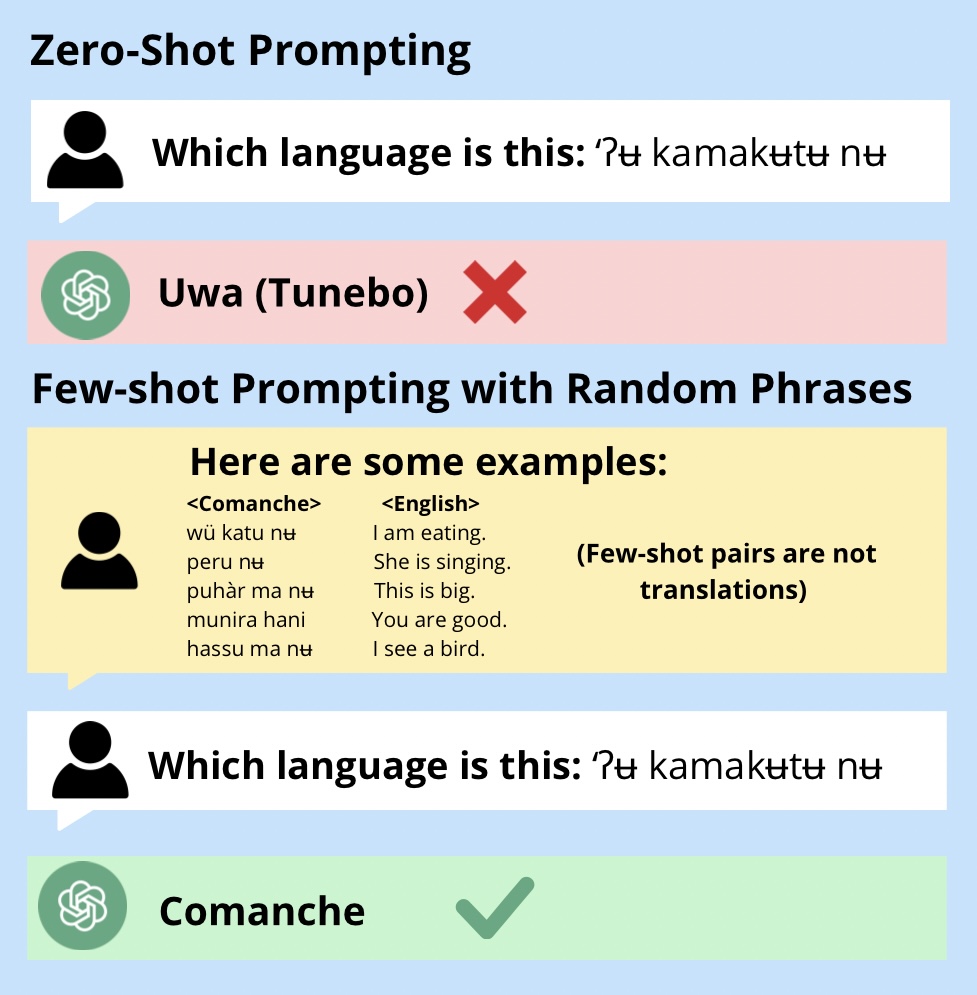}
  \caption{GPT-4o achieves a remarkable improvement in language identification performance, with  the help of few-shot examples.}
  \label{fig:langid}
\end{figure}

\section{Language Identification}
While data collection and synthetic expansion are crucial aspects of language preservation, identification is equally essential. Despite supporting over 200 languages, Google's LangID system \cite{caswell2020language} does not include a single Native American language, including Comanche, highlighting the systematic exclusion of these languages from mainstream computational resources. This absence not only limits automatic language identification capabilities, but also further marginalizes endangered languages in digital spaces, making their preservation even more challenging.

\begin{figure}[t]
\centering
\includegraphics[width=\linewidth]{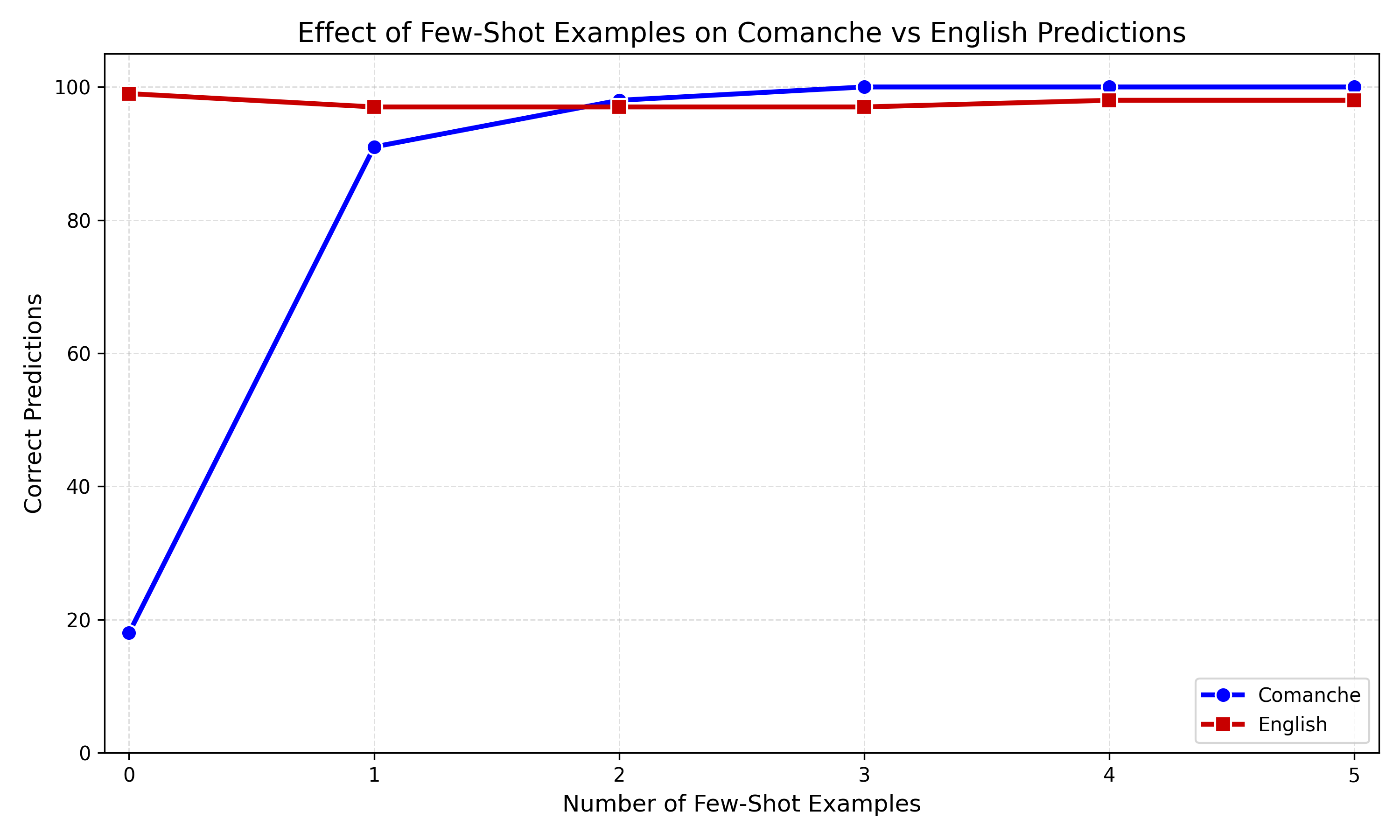}
\caption{Effect of Few-Shot Examples on Comanche Prediction Accuracy.}
\label{fig:fewshot}
\end{figure}

While LLMs have demonstrated remarkable proficiency in high-resource language tasks, their ability to identify low-resource languages remains a critical challenge. In our zero-shot prompting experiments using a dataset of 412 Comanche entries, GPT-4o achieved only 13.5\% accuracy, correctly identifying 56 instances. These results highlight the broader issue that without explicit guidance, even state-of-the-art models struggle to recognize endangered languages. To address this limitation, we introduced few-shot prompting with both Comanche and English samples, training the model to actively identify features of the Comanche language. For this experiment, we used a sample of 100 randomly selected entries from our original dataset. Each few-shot pair included one Comanche phrase and a randomized English entry from the dataset, as shown in Figure \ref{fig:langid}. With just one Comanche example, GPT-4o achieved 91\% accuracy in identification of Comanche. Extending to a three-shot strategy consistently yielded 100\% accuracy, as shown in Figure \ref{fig:fewshot}. Notably, English identification accuracy remained consistently high (97-100\%) across all experimental conditions. These findings underscore the limitations of default language identification systems and demonstrate that even minimal targeted prompting can significantly enhance recognition capabilities. The stark performance gap between Comanche and English underscores the model’s inherent bias toward high-resource languages when tasked with identification. Our results provide a scalable, low-resource approach for integrating endangered languages into NLP systems, offering a pathway toward more inclusive computational language technologies.

\section{Community Feedback}

To ensure that our approach to NLP-driven language preservation is both transparent and respectful, we engaged with a community member of Comanche and Rarámuri heritage through a semi-structured interview. The interview provided insights into the lived experiences of individuals connected to endangered languages, highlighting both the cultural significance of linguistic preservation and the challenges posed by data scarcity.

The interviewee shared that although the Comanche and Rarámuri languages were not passed down to him, he maintains a profound connection to his Native American heritage. He recounted a childhood experience in which he struggled to communicate with members of a Rarámuri community in Chihuahua, Mexico, due to language barriers\footnote{The interviewee recalled attempting to explain to local Rarámuri residents that his disposable camera differed from a Polaroid and would not produce an immediate photograph. This miscommunication left a lasting impression on him, reinforcing the importance of language technologies.}. His reflections highlight the critical role that digital resources and computational methods can play in language preservation. While exposure to artificial intelligence and NLP technologies remains limited in many Indigenous communities, the potential for these tools to support language revitalization is immense. Our study emphasizes that responsible NLP research must engage directly with affected communities, ensuring that technological interventions align with cultural needs and ethical considerations.

\section{Future Work}
Future efforts will focus on expanding the manually curated Comanche dataset, refining the synthetic data generation pipeline, and developing a real-time language identification demo. Given the largely oral nature of Comanche, we will also investigate audio-based approaches to support speech recognition and transcription, as well as exploring learning \cite{wang2019data, mangar2025engaging} and reading comprehension tasks \cite{zhang2024word}. Additionally, we will actively engage with more Comanche community members to ensure our work remains aligned with their needs and perspectives. We hope to eventually secure the resources to conduct a deeper analysis of Comanche and other indigenous languages—work that has largely been limited to high-resource languages—examining dimensions such as linguistic features \cite{lee2024llm}, implicit versus explicit expression \cite{wangimpscore}, persuasive strategies \cite{wang2024mentalmanip, yang2024enhanced} and intellectual humility \cite{guo2024computational}.

\section{Conclusion}
This study represents the first computational effort to integrate Comanche into the NLP landscape, addressing critical gaps in language documentation and technological accessibility. Through manual data collection, synthetic data expansion, and empirical evaluations of LLM-based language identification, we demonstrate that even minimal resources can yield meaningful improvements in language modeling for endangered languages. While this work marks an initial step, continued collaboration with Comanche speakers, expansion into audio-based methods, and refinement of evaluation metrics will be essential to advancing these efforts. We advocate for a NLP research paradigm that actively includes Indigenous and low-resource languages, ensuring that they are not only preserved but empowered through computational advancements.

\section*{Limitations}
Despite the contributions of this study, several limitations must be acknowledged. Firstly, the manually curated Comanche dataset remains small, constraining both model performance and generalizability. Future work must expand this dataset to improve model robustness and alignment \cite{zeng2025converging}, as well as to prevent biases \cite{guan2025saged}. In addition, while synthetic data augmentation offers a promising avenue for resource expansion, the quality of generated translations is inherently dependent on the prompting strategy \cite{jian2022contrastive} and the capabilities of the underlying language model. Further refinements to the pipeline and more rigorous evaluation methodologies are necessary to ensure linguistic accuracy. Moreover, our experiments focus primarily on text-based language identification, overlooking the oral tradition of Comanche. Future research should incorporate audio-based approaches, such as automatic speech recognition, to better align with the language’s natural form. Lastly, our engagement with community members, while valuable, represents only an initial step. Sustained collaboration with Comanche speakers and language advocates will be essential to ensuring that computational interventions align with community priorities and ethical considerations.

\section*{Ethics Statement}
Our research adheres to ethical principles that prioritize Indigenous data sovereignty, cultural sensitivity, and responsible engagement. We collected Comanche words, affixes, and phrases exclusively from publicly available sources, ensuring transparency in our data practices and proper attribution of all resources. All relevant citations for the manual dataset can be found in Appendix \ref{sec:appendixd}. Consistent with the principles outlined by \citet{schwartz-2022-primum}, we acknowledge that Indigenous languages are deeply tied to cultural identity, historical continuity, and community sovereignty. We explicitly recognize the Comanche Nation as the rightful stewards of their language and are committed to ensuring that our work aligns with their goals of preservation and revitalization. Our research seeks not only to document but to actively contribute to the accessibility and visibility of Comanche within the computational linguistics community. We emphasize the importance of relational engagement with Indigenous communities, acknowledging that linguistic data is not merely an artifact for academic study but also a living expression of cultural heritage (Appendix \ref{sec:appendixa}).

Finally, we uphold ethical obligations of cognizance, beneficence, accountability, and non-maleficence. We remain committed to avoiding harm, ensuring that our findings and datasets serve as tools for language empowerment rather than extraction. Future work will continue to involve direct engagement with Comanche speakers, fostering a collaborative research framework that respects community agency and cultural priorities. In the spirit of transparent and ethical research, our full dataset and code have been made available at (\url{https://github.com/comanchegenerate/ComancheSynthetic}).

\bibliography{anthology,custom}
\bibliographystyle{acl_natbib}

\appendix

\newpage
\section{Appendix A}
\label{sec:appendixa}

\begin{figure}[h]
\centering
\includegraphics[width=\linewidth]{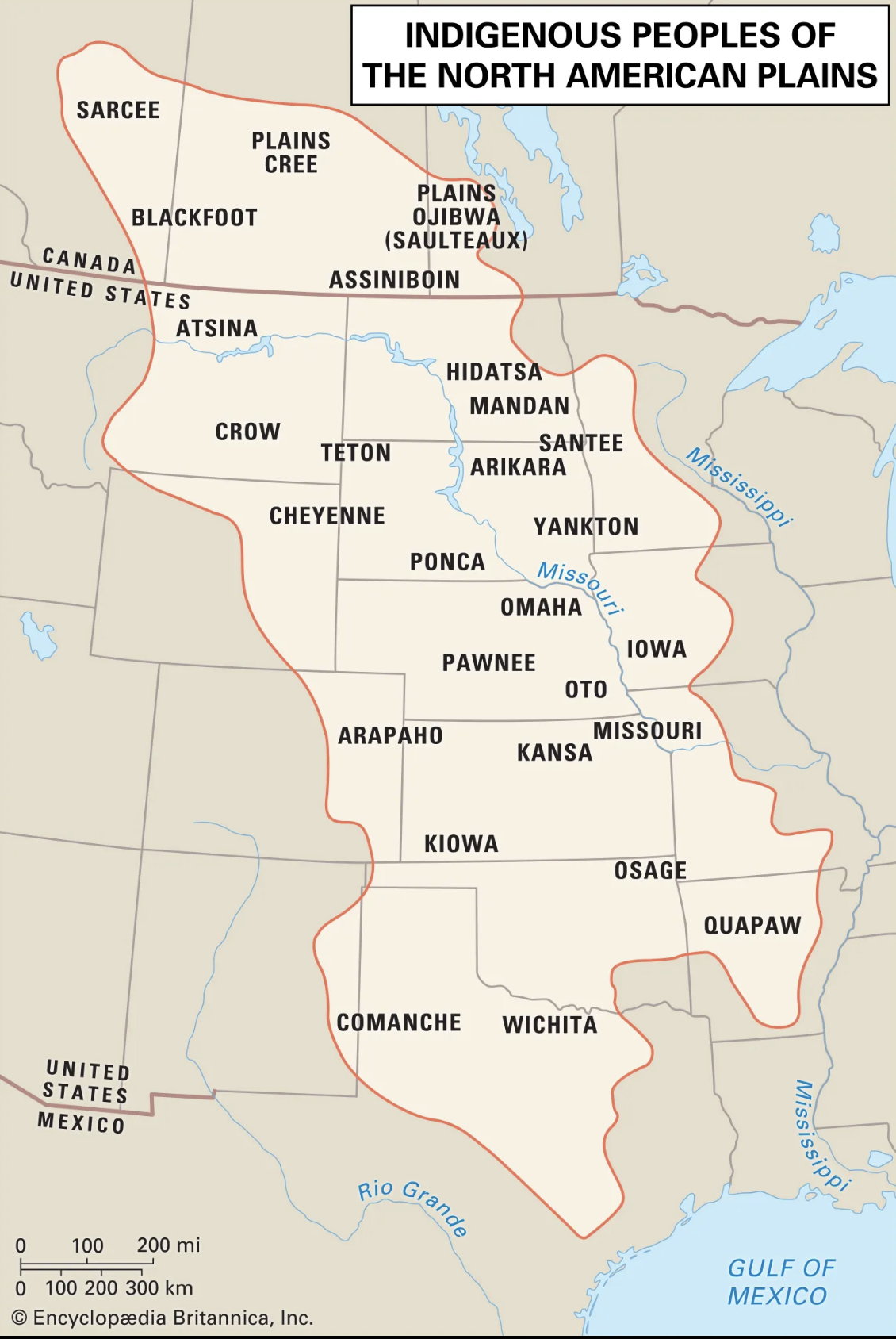}
\caption{Map showing approximate locations of Indigenous peoples of the Great Plains prior, to displacement in the 19th century. Comanche territory is depicted in the bottom-left region. Source:\url{https://www.britannica.com/place/Great-Plains\#/media/1/243562/330}.}
\label{fig:appendixb}
\end{figure}

\begin{figure}[h]
\centering
\includegraphics[width=\linewidth]{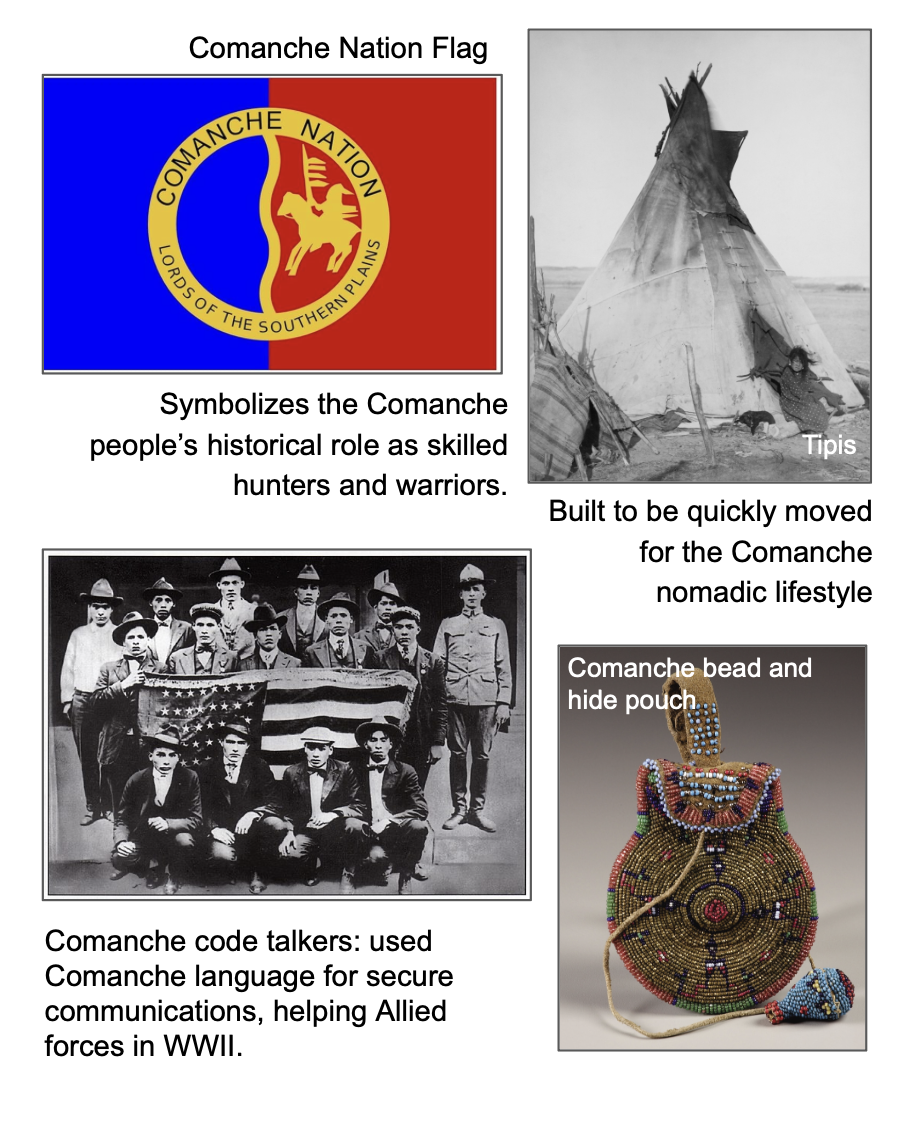}
\caption{Comanche cultural artifacts.}
\label{fig:culture}
\end{figure}

\begin{figure}[h]
\centering
\includegraphics[width=\linewidth]{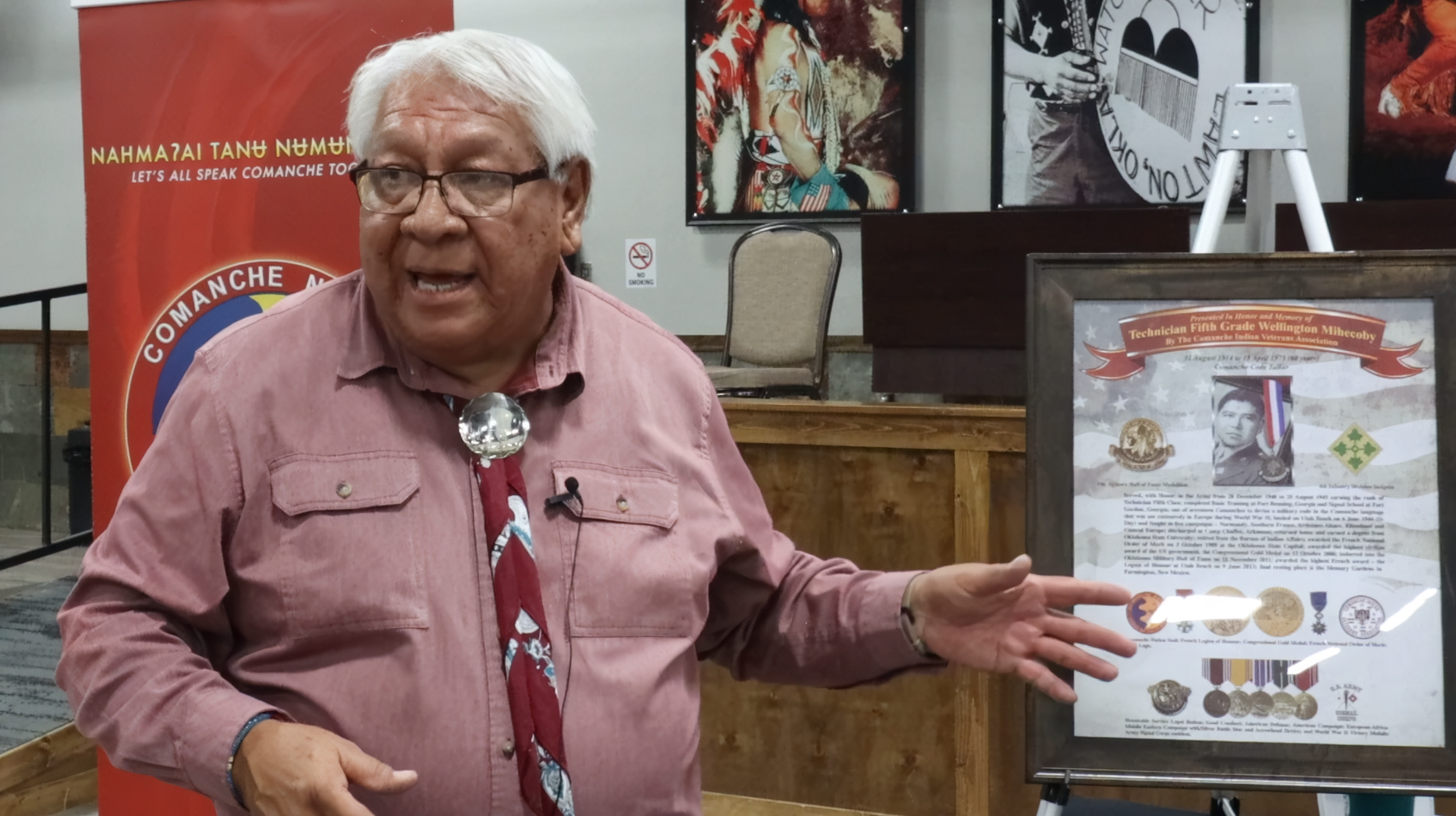}
\caption{Lloyd Heminokeky, Jr., Language Consultant for the Comanche Nation Language Department, hosts an event honoring Comanche Code Talkers, including his grandfather, Technician Fifth Grade Wellington Mihecoby, whose distinguished service is highlighted in the portrait beside him.
Source: \url{https://youtu.be/M_JO8C63Ins?si=uc8JzAiAX7sCrF9A}.}
\end{figure}

\clearpage
\section{Appendix B}
\label{sec:appendixb}

\begin{figure}[h]
\centering
\includegraphics[width=\linewidth]{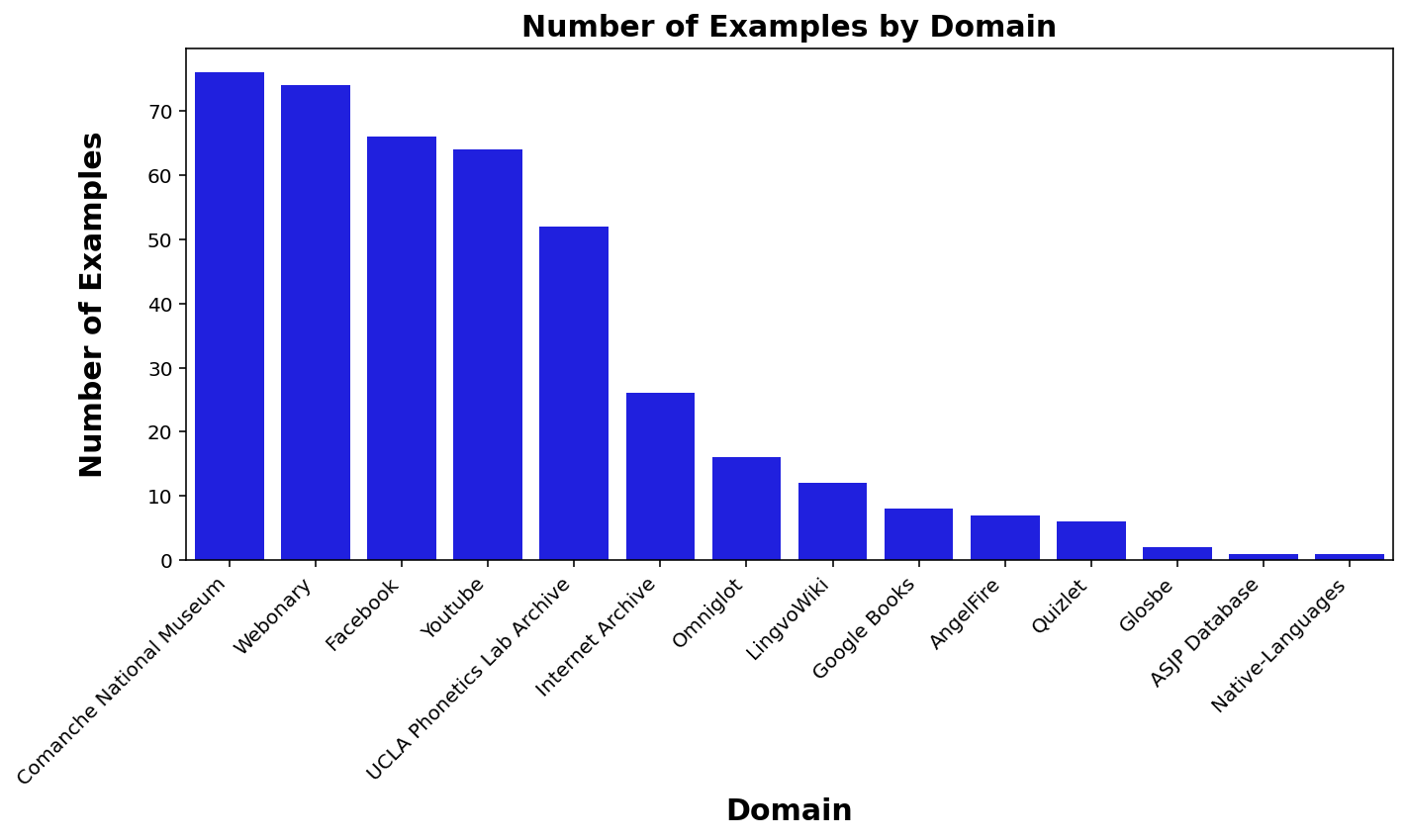}
\caption{Distribution of manually collected Comanche-English phrases across 15 sources. The Comanche National Museum, Webonary, and Facebook (via the Comanche Nation Language Department) contributed the highest number of examples. This distribution underscores the variability in available linguistic resources for Comanche.}
\label{fig:examplesdistribution}
\end{figure}

\begin{figure}[h]
    \centering
    \includegraphics[width=\linewidth]{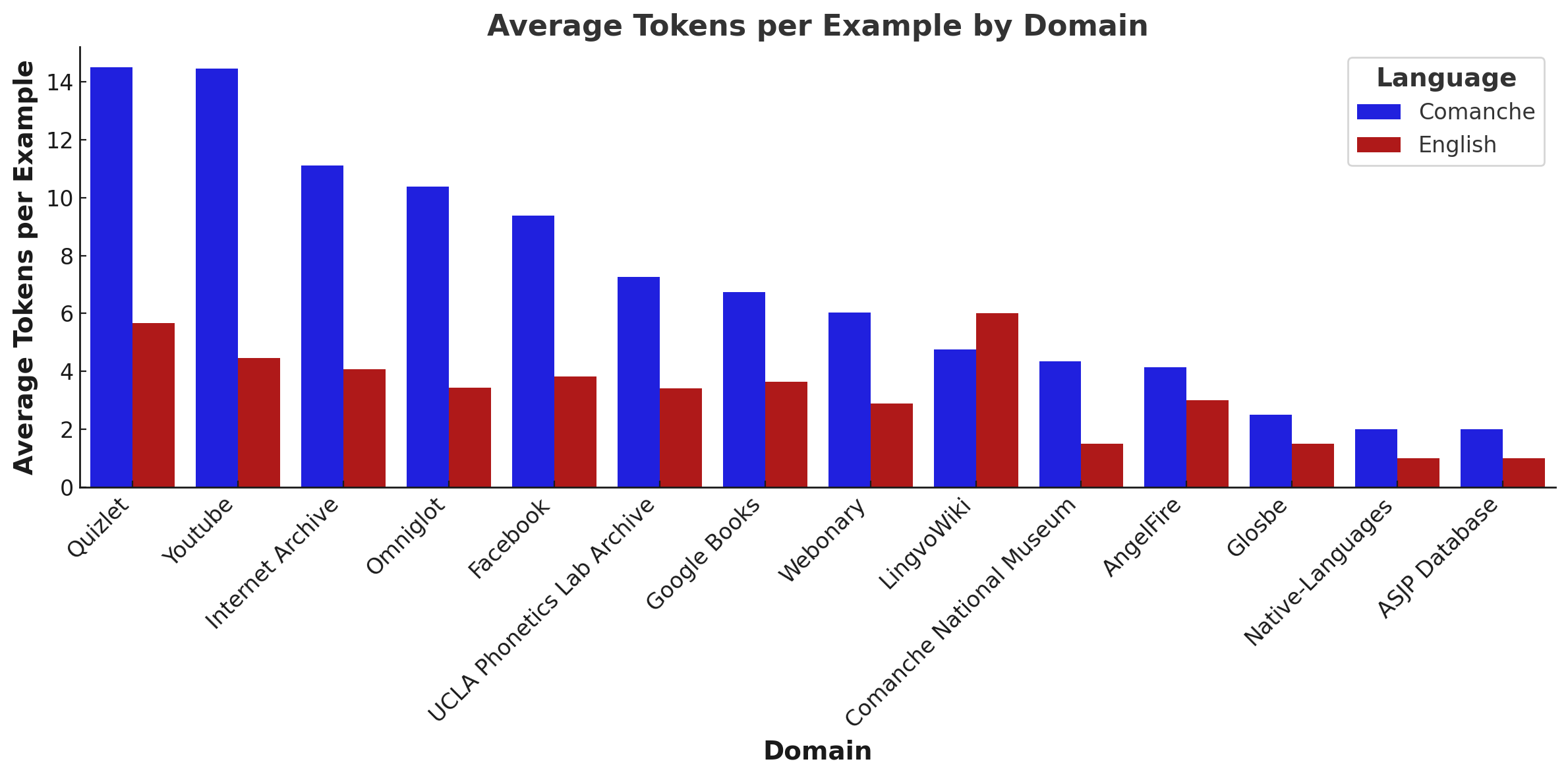}
    \caption{The average token length per example differs notably between Comanche (blue) and English (red). Some sources, such as Quizlet and Youtube, contain significantly longer Comanche phrases, while others, such as LingvoWiki, show an inverse pattern due to the presence of affixes and bound morphemes. These variations highlight source-specific differences, particularly in how morphology and translation conventions impact token length.}
    \label{fig:tokenavgdomain}
\end{figure}

\begin{figure}[h]
    \centering
    \includegraphics[width=\linewidth]{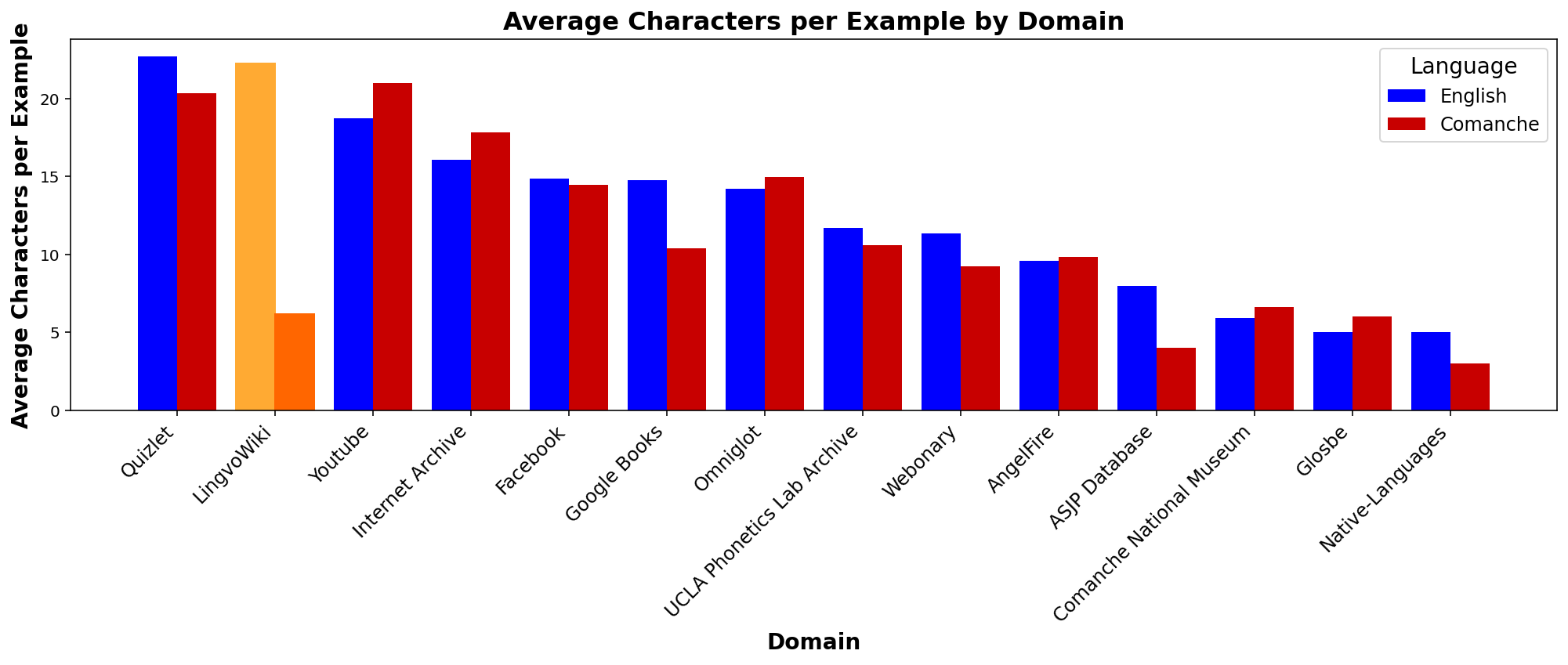}
    \caption{Comparison of average characters per example across various sources of English and Comanche. Notably, the Comanche data from LingvoWiki appears unusually short due to the presence of affixes in the collected samples, which artificially lowers the character count for that source.}
    \label{fig:avgcharbargraph}
\end{figure}

\begin{figure}[h]
    \centering
    \includegraphics[width=\linewidth]{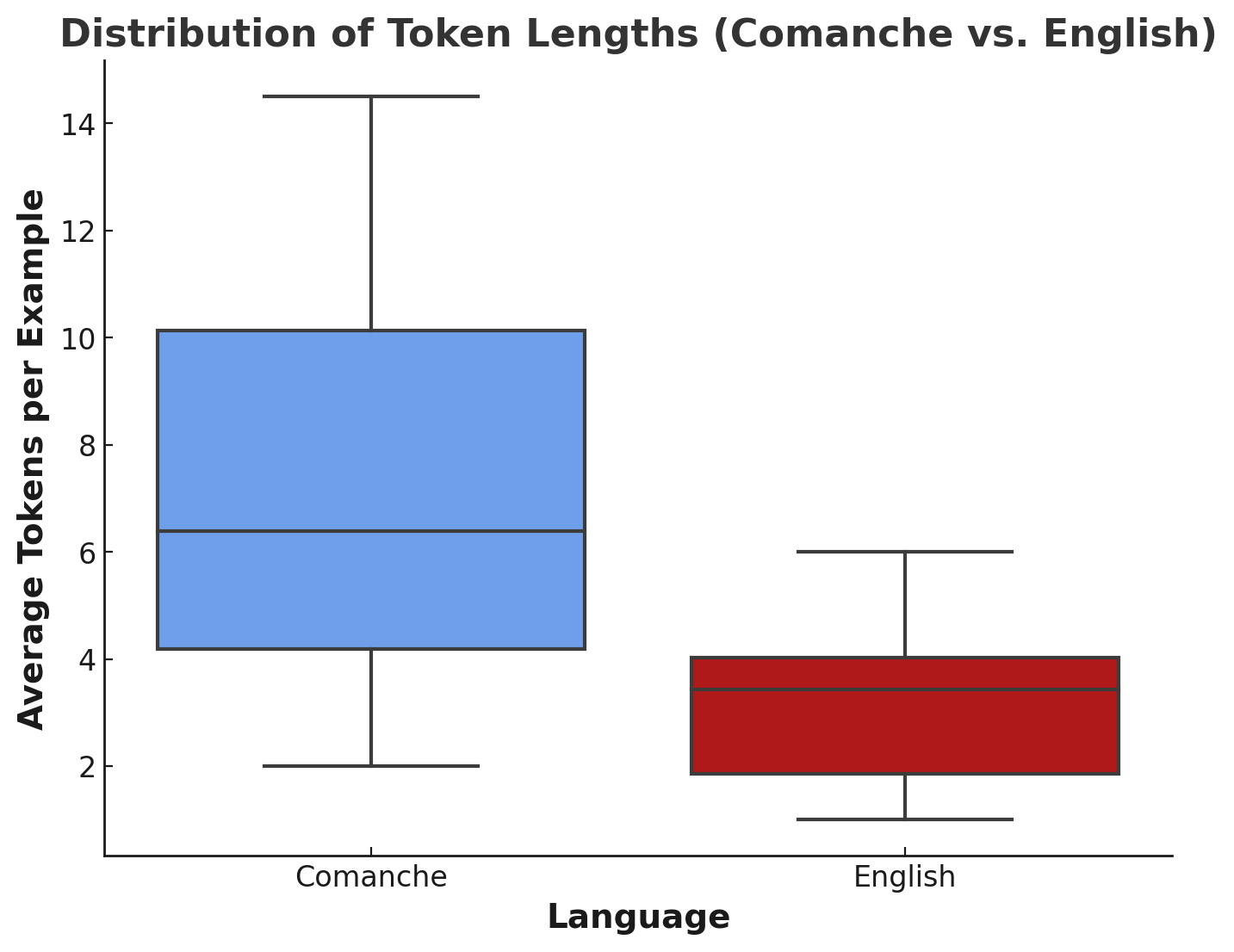}
    \caption{Box plot distribution of token lengths in Comanche and English phrases. Comanche phrases exhibit a wider range of token lengths, with a median of around 6 tokens per example, and an extended upper quartile value, reflecting its polysynthetic structure. English translations, by contrast are more compact with less variability.}
    \label{fig:examplesboxplot}
\end{figure}

\clearpage
\section{Appendix C}
\FloatBarrier
\begin{figure*}[!b]
  \centering
  \includegraphics[width=\linewidth]{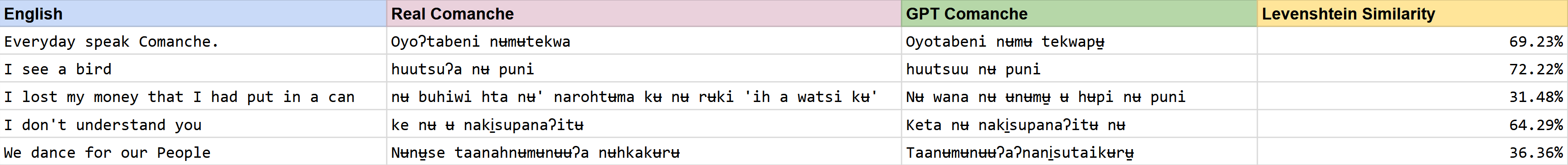}
  \caption{Comparison of English sentences with their corresponding Real Comanche translations and GPT-generated Comanche translations.}
  \label{fig:GPT-4o validation results}
\end{figure*}

\newpage
\newgeometry{left=2.5cm,right=2.5cm,top=2.5cm,bottom=2.5cm}

\section{Appendix D}
\label{sec:appendixd}

\begin{center}
\scriptsize 
\begin{longtable}{p{2.5cm} p{3.3cm} p{4.6cm} p{3.1cm}} 
\caption{Online sources referenced to construct the Comanche-English dataset.\label{tab:comanche-dataset-sources}}\\

\hline
\textbf{Cite Key} & \textbf{Author (Year)} & \textbf{Title} & \textbf{URL} \\
\hline
\endfirsthead

\hline
\textbf{Cite Key} & \textbf{Author (Year)} & \textbf{Title} & \textbf{URL} \\
\hline
\endhead

\hline
\multicolumn{4}{r}{\textit{Continued on the next page}}\\
\endfoot

\endlastfoot

\texttt{omniglotwriting} & Omniglot (n.d.) & Comanche language, alphabet and pronunciation & \url{https://www.omniglot.com/writing/comanche.htm} \\
\hline

\texttt{omniglotphrases} & Omniglot (n.d.) & Comanche phrases & \url{https://www.omniglot.com/language/phrases/comanche.htm} \\
\hline

\texttt{ucla1992} & UCLA Phonetics Lab Archive (1992) & Comanche word lists (1992) & \url{https://archive.phonetics.ucla.edu/Language/COM/} \\
\hline

\texttt{angelfire} & Angelfire (n.d.) & Comanche language page & \url{https://www.angelfire.com/creep2/fracod/comanche.html} \\
\hline

\texttt{rosettaproject} & Internet Archive (n.d.) & rosettaproject\_com\_morsyn-1 & \url{https://archive.org/details/rosettaproject_com_morsyn-1/page/n3/mode/2up} \\
\hline

\texttt{cnlanguagefb} & Comanche Nation Language Dept. (n.d.) & Facebook videos & \url{https://www.facebook.com/CNLanguage/videos/} \\
\hline

\texttt{comanchemuseum} & Comanche National Museum and Cultural Center (n.d.) & Comanche dictionary & \url{https://www.comanchemuseum.com/dictionary.html} \\
\hline

\texttt{nativelang} & Native Languages of the Americas (n.d.) & Comanche language: Word sets & \url{https://www.native-languages.org/comanche_words.htm} \\
\hline

\texttt{glosbecomanche} & Glosbe (n.d.) & English--Comanche dictionary & \url{https://glosbe.com/en/com} \\
\hline

\texttt{asjpcomanche} & ASJP (n.d.) & COMANCHE & \url{https://asjp.clld.org/languages/COMANCHE} \\
\hline

\texttt{webonarycomanche} & Comanche Dictionary Project (n.d.) & Comanche webonary & \url{https://www.webonary.org/comanche/} \\
\hline

\texttt{lingvoforum} & LingvoForum (n.d.) & Comanche dictionary (LingvoForum Wiki) & \url{https://wiki.lingvoforum.net/wiki/Comanche_dictionary} \\
\hline

\texttt{quizletcomanche} & Quizlet (n.d.) & Comanche phrases flashcards & \url{https://quizlet.com/718424723/comanche-phrases-flash-cards/} \\
\hline

\texttt{youtubecn} & Comanche Nation Language Dept. (n.d.) & CNLanguage YouTube channel & \url{https://www.youtube.com/@CNLanguage/videos} \\
\hline

\end{longtable}
\end{center}

\restoregeometry

\end{document}